\documentclass[11pt]{article}
\usepackage[T1]{fontenc}
\usepackage[utf8]{inputenc}
\usepackage[margin=1in]{geometry}
\usepackage{newtxtext,newtxmath}
\usepackage{microtype}
\usepackage{graphicx}
\usepackage{booktabs}
\usepackage{amsmath}
\usepackage[numbers,sort&compress]{natbib}
\usepackage{authblk}
\usepackage{caption}
\usepackage[hidelinks]{hyperref}
\usepackage{url}
\title{What Does 4-Bit Microscaling Cost a Wearable Activity Recognizer?\\[0.4em]
\large MXFP4 versus NVFP4 on Hang-Time HAR, Evaluated Player by Player}
\author{Megan Healy}
\author{Hossein Valavi}
\affil{Princeton University, Princeton, NJ, USA\protect\\
\texttt{mh3265@princeton.edu}, \texttt{hvalavi@princeton.edu}}
\date{}

\begin{document}
\maketitle

\begin{abstract}
Wearable sports sensors can recognize actions on the device itself, where lower-precision numbers reduce model storage and memory traffic.
Two 4-bit block-scaled formats, MXFP4 from the Open Compute Project, introduced in 2023, and NVFP4 from NVIDIA, formally introduced in 2025, are drawing growing interest. Published evaluations cover general deep-learning benchmarks and large language models; to our knowledge, nobody has yet measured what they cost a small wearable-sensor model.
We train a 159,492-parameter 1D CNN to recognize four basketball actions from wrist-accelerometer data in Hang-Time HAR and test it on players it has never seen, using five-fold cross-validation grouped by player. Every comparison between number formats is made fold by fold.
In full precision the model reaches 69.26\% macro-F1 (the mean of the four per-class F1 scores). Measured TFLite FP16 and INT8 models lose at most 0.16 percentage points. In simulation, 4-bit weights cost 1.45 points with MXFP4 and 0.72 points with NVFP4. When activations are quantized to 4 bits as well, MXFP4 loses 5.14 points and NVFP4 only 0.75, and NVFP4 comes out ahead in all five held-out-player folds.
The ranking survives two alternative simulator settings, with NVFP4 leading by 3.49 to 4.40 points. A $2 \times 3$ ablation with 4-bit weights and activations (two block sizes and three scale rules, with the same 4-bit elements throughout) points to scale selection as the main source of MXFP4's loss. The reference MXFP4 conversion that we simulate rounds the scale down, which can pick scales that are too small, so many blocks clip their largest values. Rounding the scale up instead recovers 3.18 points with 32-value blocks and 3.63 points with 16-value blocks, and switching to an 8-bit floating-point scale, as NVFP4 does, adds another 0.84 and 1.20 points.
Block size matters less. Under this round-down rule, 16-value blocks score 0.44 points below 32-value blocks in all five folds; under the other two rules there is no consistent effect.
\end{abstract}


\section{Introduction}
Players in team sports can wear a small inertial measurement unit (IMU) on the wrist or, in a sport like hockey, mount one on the stick.
If the sensor classifies actions itself, it can give feedback during play while saving battery and radio bandwidth.
The catch is memory. Small devices have very little of it, so the model has to be stored, and ideally computed, in few bits.

On edge devices INT8 is the established choice: weights, and optionally activations, are stored as 8-bit integers together with a scale~\cite{jacob2018quant,tfliteptq}.
More recently, 4-bit \emph{block-scaled} floating-point formats have appeared.
MXFP4, defined by the Open Compute Project (OCP) in 2023, stores each number as a 4-bit float (E2M1) and lets a block of 32 numbers share one power-of-two scale~\cite{ocp2023mx,rouhani2023mx}.
NVFP4, introduced by NVIDIA in 2025, uses blocks of 16, an 8-bit floating-point (FP8 E4M3) scale per block, and one 32-bit scale per tensor~\cite{nvidia2025nvfp4}.
Both formats are aimed at deep-learning accelerators~\cite{rouhani2023mx,nvidia2025nvfp4}. Evaluations so far cover a broad range of deep-learning benchmarks~\cite{rouhani2023mx} and large-language-model training~\cite{yang2025mxstudy}.
Wearable activity-recognition CNNs process low-rate sensor signals and can be very small, so their weight and activation statistics may differ. We are not aware of any measurement of what these formats cost such models.

Our question is therefore: \emph{how much accuracy do MXFP4 and NVFP4 cost a small wearable activity recognizer, compared with FP32, FP16 and INT8, and which design choice---block size or scale type---matters?}
We use basketball data from Hang-Time Human Activity Recognition (HAR)~\cite{hoelzemann2023hangtime} as a stand-in for our target sport, ice hockey.
An ice-hockey dataset does exist: Vleugels et al.\ recorded six activities with a stick-worn IMU from five players and describe their data set as open source~\cite{vleugels2021hockey}.
We use Hang-Time HAR instead because our evaluation holds out whole players: with only five players, a five-fold player-grouped split would score each fold on a single test player, and the hockey recordings total about 72 minutes~\cite{vleugels2021hockey}, whereas Hang-Time HAR's 24 players give 4--5 test players per fold.

This paper makes three contributions:
\begin{itemize}
\item We compare FP32, FP16, INT8, INT4, MXFP4 and NVFP4 on one small HAR CNN, fold by fold (paired) with player-grouped folds, quantizing either the weights only (W4) or both weights and activations (W4A4).
\item A 2$\times$3 ablation study (block size $\times$ scale rule) separates the design differences between MXFP4 and NVFP4. In this setting, the round-down scale rule of the reference MXFP4 conversion, which we use as our MXFP4 baseline, explains most of its loss, in line with earlier LLM-training results on scale rounding~\cite{yang2025mxstudy}.
\item We state plainly what the evidence does and does not support: per-fold results, which differences hold in every fold, two sensitivity checks of the simulator (tie rounding and first-layer blocking), and where simulation may be optimistic.
\end{itemize}
The FP32, FP16 and INT8 results are \emph{measured} on real TFLite files.
All 4-bit results are \emph{simulated} (fake quantization) and come without speed or energy numbers, since the CPUs we used have no hardware for 4-bit floating-point arithmetic; measurements on FP4-capable accelerators~\cite{nvidia2025nvfp4} are left for future work.

\section{Background and Related Work}
\textbf{Number formats.}
Names such as E2M1, E4M3 and E8M0 give the number of exponent bits (after E) and mantissa bits (after M).
FP4 (E2M1) has one sign bit, two exponent bits and one mantissa bit, so it can only represent the magnitudes 0, 0.5, 1, 1.5, 2, 3, 4 and 6, with either sign~\cite{ocp2023mx}.
What makes such a coarse format usable is block scaling: every block of values gets its own scale.
In MXFP4 the shared scale is $2^{\lfloor \log_2 \max_i |v_i| \rfloor - 2}$, where 2 is FP4's largest exponent; elements that overflow are clamped to $\pm 6$~\cite{ocp2023mx,rouhani2023mx}.
Because the scale exponent is rounded \emph{down}, a block's largest values can exceed 6 and get clipped.
The specification requires implementations to support round-to-nearest, ties-to-even when converting to FP4, but it also permits other conversion recipes~\cite{ocp2023mx,rouhani2023mx}.
NVFP4 takes a different approach. It sets each block's scale to the block's largest absolute value divided by 6, so that the largest magnitude lands exactly on 6, the largest FP4 value.
This scale is stored as an FP8 E4M3 number~\cite{micikevicius2022fp8} and multiplied by one FP32 scale per tensor~\cite{nvidia2025nvfp4}.
Including the block scales, MXFP4 costs 4.25 bits per value and NVFP4 4.5; NVFP4's FP32 tensor scale adds $32/N$ bits per value for a tensor of $N$ values.
Yang et al.\ compared scale rounding modes for MX formats in LLM training~\cite{yang2025mxstudy}; we ask the same question for a small sensor CNN at inference time.

\textbf{Sports HAR.}
For ice hockey, Vleugels et al.\ classified six activities from a stick-worn IMU with a CNN and reached 76\% accuracy on an unseen player~\cite{vleugels2021hockey}.

\textbf{Quantized HAR.}
Daghero et al.\ deploy 1D CNNs for HAR on microcontrollers using sub-byte and mixed-precision integer quantization~\cite{daghero2022har}.
We know of no work that evaluates MXFP4 or NVFP4 on wearable HAR.

\textbf{Evaluation.}
Consecutive sliding windows overlap (by 80\% in this study; Section~\ref{sec:method}), so a random split places near-copies of test windows in the training set.
Subject-dependent evaluation can therefore inflate results~\cite{dehghani2019windows}, and subject-independent schemes such as leave-one-subject-out are recommended instead~\cite{gholamian2020loso}.
Because cross-validation folds share training data, there is no unbiased estimator of the variance of a K-fold result~\cite{bengio2004cv}.
For this reason we report paired per-fold differences rather than comparing overlapping error bars.

\section{Method}\label{sec:method}
\textbf{Data.}
Hang-Time HAR contains wrist accelerometer recordings (about 50\,Hz) of 24 basketball players from two teams, labeled from video~\cite{hoelzemann2023hangtime}.
We use the three acceleration channels and four ball actions: dribbling, shot, pass and layup.
Players are identified by recording file, since the same 4-character code can appear in both teams.
We cut 1\,s windows (50 samples) and slide the window forward by 0.2\,s (10 samples) at a time.
This yields five windows per second of recording, and consecutive windows share 0.8\,s, or 80\%, of their samples.
The window length follows from the actions themselves. A window should be shorter than what it has to recognize, and the shortest action, the pass, has a median duration of 1.20\,s in our measurements, so a 1\,s window fits inside a typical action.
The 0.2\,s step is a fixed setting of the pipeline that we did not vary.
Each window takes the label of its center sample and is kept only if at least half of its samples share that label, which gives 118,823 windows.
Left unbalanced, the data would let a model score well by mostly predicting dribbling.
We therefore cap dribbling at three times the shot count (fixed seed), which leaves 75,865 windows: 45,699 dribbling, 15,233 shot, 8,657 pass and 6,276 layup.
The cap reduces the dominance of dribbling while keeping every window of the other three actions; its factor of 3 is a fixed design choice that we did not vary.
Since the cap is applied before splitting, the test folds are capped too, and dribbling falls from 74.61\% to 60.24\% of windows.
Absolute scores thus refer to this capped distribution, and every format is scored on the same windows.
Note also that the task is to classify windows already known to contain one of the four actions; we do not detect actions against background activity.

\textbf{Evaluation.}
We use 5-fold \texttt{GroupKFold} grouped by player, so each player is tested exactly once, by a model that has never seen them (4--5 players per fold).
Our score is macro-F1, the mean of the four per-class F1 scores, reported on a 0--1 scale; a change of 0.01 is one percentage point, the unit used in the abstract.
For reference, always answering ``dribbling'' gives macro-F1 0.188.
Every format is evaluated with the same five trained models on the same five test folds.
We compare formats by their \emph{paired} per-fold difference and call a difference consistent only if it has the same sign in all five folds.
With five folds this is the strongest result we can get, but it is not proof: if a format had no effect, all five folds would still agree in sign with probability $2/2^5 = 0.0625$.
Where we report a spread, it is the population standard deviation (sd) over the five folds.

\textbf{Model.}
The 1D CNN has three convolution blocks with 64, 128 and 256 filters. In each block the convolution is followed by ReLU, batch normalization, max-pooling and dropout, in that order.
Global average pooling, a 128-unit dense layer and a 4-way softmax complete the network.
In total the model holds 159,492 values: 158,596 trainable parameters plus 896 batch-normalization running statistics (each channel's running mean and variance, which are updated during training but not learned by the optimizer).
Each fold's model is trained with Adam (batch 256, up to 25 epochs).
Early stopping (patience 5) monitors about 20\% of the \emph{training} players and never the test players, because using test players here would leak test information into model selection.

\textbf{Deployable formats (measured).}
We convert each fold's model to TFLite in four ways: FP32; FP16 weights; dynamic-range INT8, which stores weights as INT8 and keeps activations in float, although kernels that support it quantize activations to 8 bits on the fly~\cite{tfliteptq}; and full-integer INT8, which also puts activations in INT8, with ranges calibrated on 500 training windows~\cite{tfliteptq}.
All files run in the TFLite interpreter of TensorFlow 2.21.0.

\textbf{4-bit formats (simulated).}
We round numbers to each format and back to FP32 (fake quantization) and then run the network as usual.
Simulation is necessary because the CPUs we used have no 4-bit floating-point arithmetic. Fake quantization reproduces the rounding error of a 4-bit format while the arithmetic itself stays in FP32, so it captures accuracy and storage but not speed or energy.
The weights of every Conv1D and Dense layer, including the first and the last, are blocked along each output channel's inputs, which is the direction of the multiply-add.
Biases and batch-normalization parameters stay in FP32.
For W4A4 we also quantize the input of every Conv1D and Dense layer, in blocks along the channel axis at each time step, with activation scales computed afresh for every window.
Each value is rounded to the nearest FP4 value. When a value lies exactly halfway between two FP4 values (for example 0.75, between 0.5 and 1), we keep the smaller magnitude. The round-half-to-even rule, which the OCP specification requires hardware to support~\cite{ocp2023mx}, would instead pick the neighbor whose last bit is 0. As Section~\ref{sec:sens} shows, this choice changes almost nothing.
In the first convolution, each time step's three input channels form their own activation block, whereas each weight block spans the whole 15-value kernel patch (5 time steps, the kernel's ``taps'', $\times$ 3 channels).
A convolution lowered to a matrix product would instead share one activation scale per 15-value patch; Section~\ref{sec:sens} tests both options.
We simulate MXFP4 (block 32, E8M0 scale rounded down, the reference conversion of the specification~\cite{ocp2023mx}) and NVFP4 (block 16, E4M3 block scale times a per-tensor FP32 scale of $\max|x|/(6\times 448)$~\cite{nvidia2025nvfp4}).
For reference we also simulate symmetric INT8 and INT4.
These use one scale per output channel for weights and \emph{one scale per window} for activations, which is much coarser than the 16- or 32-value blocks of the FP4 formats.
Finally, the ablation study (called Experiment C in the accompanying notebook) crosses block size $\{32, 16\}$ with three scale rules: E8M0 rounded down (MX rule), E8M0 rounded up (no clipping), and E4M3 with a tensor scale (NV rule).

\textbf{Self-tests and provenance.}
With nothing rounded, our layer-by-layer forward pass reproduces the trained models' predictions exactly.
A second forward pass, written independently in NumPy without TensorFlow, reproduces the simulated predictions exactly for FP32 and MXFP4, and for 75,864 of 75,865 windows for NVFP4.
These checks show that the code does what we describe; they do not show that it matches FP4 hardware.
As a further check at 8 bits, simulated W8A8 gives macro-F1 0.6930, against 0.6910 for the real full-integer TFLite model.
Finally, the models' saved optimizer step counts are exact multiples of the batches per epoch on the inner-training players (12, 16, 16, 13 and 7 epochs), which confirms that early stopping used the inner validation players.

\textbf{Use of generative AI.}
The research question, data, model, training setup, number formats and the $2\times3$ ablation were defined by the authors in their original analysis notebook.
Claude (Anthropic), a generative AI assistant, wrote additional analysis and verification code: the independent NumPy forward pass, the checkpoint-provenance check, the paired per-fold statistics, the two sensitivity reruns of Section~\ref{sec:sens}, a checklist that compares every number reported here with the notebook's output, and the scripts that produce the figures and tables.
The authors ran this code, inspected its outputs, and take responsibility for all results and interpretations.

\section{Results}
\textbf{Baselines.}
The CNN reaches macro-F1 0.6926\,$\pm$\,0.0494 (standard deviation over five folds) on unseen players. Its per-fold scores are 0.660, 0.660, 0.765, 0.738 and 0.641.
It beats a random forest trained on 18 hand-crafted window features (0.608, pooled) in all five folds, by $+$0.0841 on average.
Splitting windows at random rather than by player raises the forest's pooled macro-F1 from 0.608 to 0.753, a clear example of the leakage described in~\cite{dehghani2019windows}.

\textbf{Deployable formats (Table~\ref{tab:tflite}).}
FP16 and dynamic-range INT8 change macro-F1 by at most 0.0003, and the sign of the change differs between folds.
Full-integer INT8 is lower in all five folds, by 0.0016 on average, and agrees with FP32 on 98.68\% of windows, while the file shrinks from 630.9\,KiB to 182.8\,KiB.
On our CPU, median single-window latencies range from 0.0807 to 0.0867\,ms.
That spread (7.4\%) is smaller than the change we saw between two sessions on the same type of Colab virtual machine (VM): an earlier run of the same files measured 0.0696--0.0775\,ms, 11.6--24.6\% lower. We therefore make no speed claim.

\begin{table}[t]
\caption{Deployable TFLite versions (\emph{measured}). Macro-F1 is mean\,$\pm$\,sd over 5 player-grouped folds. $\Delta$ is the paired mean difference from FP32, with the number of folds in which the version scored lower. Latency: median of 5 interleaved rounds of 2,000 single-window runs, 1 thread, on a Google Colab CPU virtual machine (VM; Intel Xeon @ 2.20\,GHz, KVM, 2 vCPUs = 1 core $\times$ 2 threads; Ubuntu 24.04.5; Python 3.13.16; TensorFlow 2.21.0). Latencies of the same files differed by 11.6--24.6\% between two Colab sessions, so these differences are within noise.}
\label{tab:tflite}
\small
\begin{tabular}{lrcccr}
\toprule
Version & File (KiB) & Macro-F1 & \begin{tabular}[b]{@{}c@{}}$\Delta$ vs.\ FP32\\(folds lower)\end{tabular} & \begin{tabular}[b]{@{}c@{}}Same answer\\as FP32 (\%)\end{tabular} & \begin{tabular}[b]{@{}c@{}}p50 latency\\(ms)\end{tabular} \\
\midrule
FP32 & 630.9 & 0.6926\,$\pm$\,0.0494 & 0 (identical) & 100.00 & 0.0867 \\
FP16 & 323.8 & 0.6925\,$\pm$\,0.0494 & $-$0.0001 (3/5 lower) & 99.99 & 0.0866 \\
INT8, dynamic range & 181.0 & 0.6924\,$\pm$\,0.0494 & $-$0.0003 (3/5 lower) & 99.81 & 0.0807 \\
INT8, full integer & 182.8 & 0.6910\,$\pm$\,0.0495 & $-$0.0016 (5/5 lower) & 98.68 & 0.0865 \\
\bottomrule
\end{tabular}
\end{table}

\begin{table}[t]
\caption{Simulated formats (fake quantization; no speed or energy). Bits/weight include scale storage: per-channel FP32 scales for INT8 and INT4, block scales for the FP4 formats, and NVFP4's FP32 tensor scale. Weights KiB is calculated for the quantized kernels plus FP32 biases and batch-normalization parameters. $\Delta$ is the paired mean difference from FP32 (0.6926), with the number of folds (of 5) in which the format scored lower. INT4 and INT8 use one scale per window for activations, so their W4A4/W8A8 rows are not like-for-like with the blocked FP4 formats.}
\label{tab:sim}
\small
\begin{tabular}{lrrcccc}
\toprule
 & & & \multicolumn{2}{c}{Weights only (W4 / W8)} & \multicolumn{2}{c}{Weights + activations (W4A4 / W8A8)} \\
\cmidrule(lr){4-5}\cmidrule(lr){6-7}
Format & Bits/weight & Weights KiB & Macro-F1 & $\Delta$ (folds lower) & Macro-F1 & $\Delta$ (folds lower) \\
\midrule
INT8 & 8.12 & 165.0 & 0.6928 & $+$0.0002 (1/5) & 0.6930 & $+$0.0004 (2/5) \\
INT4 & 4.12 & 88.2 & 0.6921 & $-$0.0006 (2/5) & 0.5493 & $-$0.1433 (5/5) \\
MXFP4 & 4.25 & 90.8 & 0.6782 & $-$0.0145 (5/5) & 0.6412 & $-$0.0514 (5/5) \\
NVFP4 & 4.50 & 95.6 & 0.6854 & $-$0.0072 (5/5) & 0.6851 & $-$0.0075 (4/5) \\
\bottomrule
\end{tabular}
\end{table}

\begin{figure}[t]
\centering
\includegraphics[width=\textwidth]{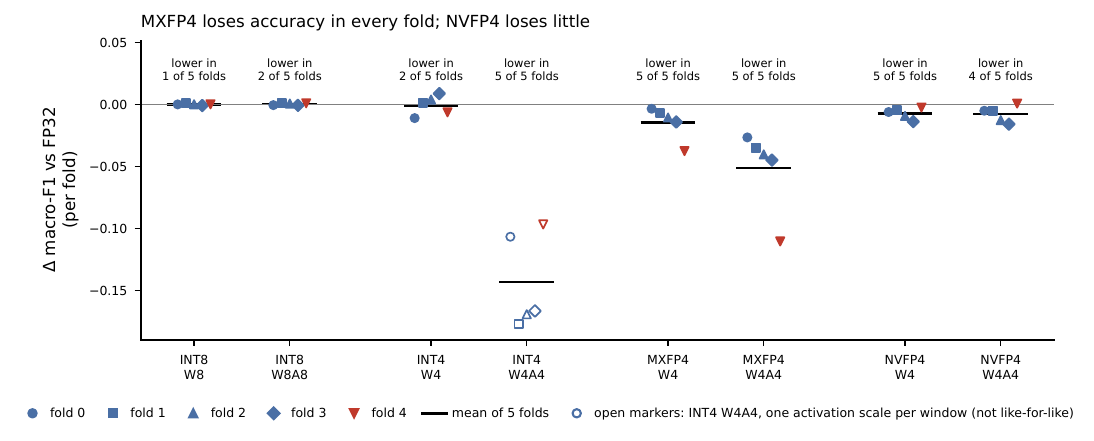}
\caption{Change in macro-F1 from FP32 for every fold (markers) and the mean of the 5 folds (black line), \emph{simulated}. The text above each group gives the number of folds (groups of test players) in which that format scored lower than FP32. Fold 4 (red) accounts for most of MXFP4's mean loss. Open markers: INT4 W4A4 uses one activation scale per window and is not like-for-like with the blocked FP4 formats. $n = 5$ folds; the same 5 trained models throughout.}
\label{fig:perfold}
\end{figure}

\textbf{4-bit weights (Table~\ref{tab:sim}, Figure~\ref{fig:perfold}).}
With weights only, MXFP4 and NVFP4 are both lower than FP32 in all five folds, by 0.0145 and 0.0072 on average.
INT4 with per-channel scales ($-$0.0006) changes direction from fold to fold.
NVFP4 beats MXFP4 in only four of five folds, so the two formats do not differ consistently when only the weights are quantized.

\textbf{4-bit weights and activations.}
Quantizing the activations as well widens the gap.
MXFP4 drops by 0.0514, is lower in all five folds, and changes the prediction for 7.29\% of windows; NVFP4 changes 3.12\%.
NVFP4's mean change ($-$0.0075) comes from four lower folds and one slightly higher fold, so it is not a consistent loss.
NVFP4 is higher than MXFP4 in all five folds, by 0.0439 on average.
MXFP4 hurts the rare actions most: layup F1 falls from 0.539 to 0.455 and pass F1 from 0.582 to 0.516, whereas NVFP4 keeps them at 0.524 and 0.575.
INT4's large drop ($-$0.1433) mainly reflects its single activation scale per window rather than the INT4 number format itself.

\begin{figure}[t]
\centering
\includegraphics[width=0.62\textwidth]{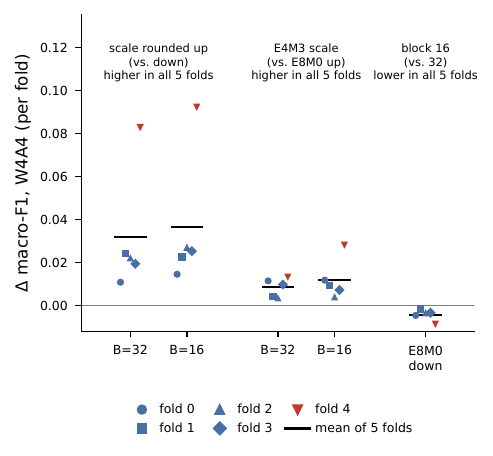}
\caption{Ablation study, W4A4, \emph{simulated}: per-fold change in macro-F1 from each design step (markers; fold 4 red; black line = mean of 5 folds). B is the block size. The text above each group gives the number of folds in which the step helped or hurt. Rounding the E8M0 scale up gives most of the gain; an E4M3 scale adds a smaller, also consistent, gain. Under the MX rule (E8M0 rounded down), 16-value blocks are slightly worse than 32-value blocks.}
\label{fig:design}
\end{figure}

\textbf{Why NVFP4 does better (Table~\ref{tab:design}, Figure~\ref{fig:design}).}
Replacing the MX rule of our MXFP4 baseline with the NV rule improves W4A4 macro-F1 by 0.0442 on average, consistently across all five folds.
We can split this improvement into two steps.
The first step rounds the power-of-two scale \emph{up} instead of down, which gains $+$0.0318 (block 32) and $+$0.0363 (block 16), again in all five folds.
The two rules pick different scales only in blocks whose largest value would exceed 6 under round-down. In those blocks, rounding up removes the clipping but doubles the step size, so this gain is the net effect of handling those blocks differently.
The second step replaces the rounded-up power-of-two scale with an E4M3 scale and gains a further $+$0.0084 and $+$0.0120, also in all five folds.
Thus 79.1\% (block 32) and 75.1\% (block 16) of the gap between the MX rule and the E4M3 rule comes from how the clipped blocks are handled (our calculation from the mean gains).
Clipping is common: in the fold-0 model, the MX rule clips the largest weight in 2,129 of 4,944 weight blocks (43\%).
If the leading digits of block maxima were spread evenly on a log scale, the expected share would be $\log_2(4/3) \approx 41.5\%$ (our calculation).
Block size has no consistent effect under the up-rounded or E4M3 rules.
Under the MX rule, however, block 16 is \emph{worse} than block 32 in all five folds ($-$0.0044).
One plausible explanation, which we have not tested, is that twice as many blocks means twice as many clipped maxima.
Because the ablation compares six prespecified variants on the same test folds, we treat it as exploratory.
At MXFP4's storage cost (4.25 bits), block-32 E4M3 scales give $-$0.0112.
Compared with NVFP4 at 4.5 bits, this variant is higher in two folds and lower in three, so the extra 0.25 bits per weight that NVFP4 spends buy no consistent gain here.

\begin{table}[t]
\caption{Ablation study (\emph{simulated}): mean change in macro-F1 from FP32, weights only / weights + activations. MXFP4 = block 32 with E8M0 rounded down; NVFP4 = block 16 with E4M3.}
\label{tab:design}
\small
\begin{tabular}{lccc}
\toprule
Block & E8M0 down & E8M0 up & E4M3 + tensor \\
\midrule
32 & $-$0.0145 / $-$0.0514 & $-$0.0063 / $-$0.0196 & $-$0.0080 / $-$0.0112 \\
16 & $-$0.0167 / $-$0.0558 & $-$0.0073 / $-$0.0195 & $-$0.0072 / $-$0.0075 \\
\bottomrule
\end{tabular}
\end{table}

\subsection{Sensitivity to simulator choices}\label{sec:sens}
We reran the two headline W4A4 formats with two changes, one at a time (Table~\ref{tab:sens}).
Both are sensitivity checks only; the main tables keep the notebook's definitions.
\emph{Ties to even.} Exact ties are rare and occur almost only in the first layer's input, where the accelerometer values lie on a coarse grid (12,702 of 11,379,750 input values for MXFP4, 0.11\%).
Rounding them to even changes 31 MXFP4 predictions and 1 NVFP4 prediction, and no mean moves by more than 0.0001.
No NVFP4 block scale underflowed to zero.
\emph{Patch blocking.} Sharing one activation scale per 15-value first-layer patch makes MXFP4 better by 0.0069 (higher in all five folds) and NVFP4 worse by 0.0020.
With patch blocking, NVFP4's W4A4 loss becomes consistent (lower than FP32 in all five folds), and NVFP4 still beats MXFP4 in all five folds.
That MXFP4 improves with larger blocks fits the clipping explanation, since fewer blocks means fewer clipped maxima.

\begin{table}[t]
\caption{Sensitivity of W4A4 results (\emph{simulated}). Change in macro-F1 from FP32, with folds lower (of 5); last column: NVFP4 minus MXFP4, with folds in which NVFP4 is higher. Main: the notebook's definitions (ties to smaller; one layer-1 activation block per time step).}
\label{tab:sens}
\small
\setlength{\tabcolsep}{3pt}
\begin{tabular}{lccc}
\toprule
Variant & MXFP4 & NVFP4 & NVFP4 $-$ MXFP4 \\
\midrule
Main & $-$0.0514 (5/5) & $-$0.0075 (4/5) & $+$0.0439 (5/5) \\
Ties to even & $-$0.0514 (5/5) & $-$0.0075 (4/5) & $+$0.0440 (5/5) \\
Layer-1 patch blocks & $-$0.0444 (5/5) & $-$0.0095 (5/5) & $+$0.0349 (5/5) \\
\bottomrule
\end{tabular}
\end{table}

\section{Discussion and Limitations}
\textbf{One fold dominates the mean.}
MXFP4's largest losses occur in fold 4: $-$0.0376 (W4) and $-$0.1105 (W4A4), compared with $-$0.0034 to $-$0.0140 and $-$0.0265 to $-$0.0448 in the other folds.
Fold 4's model also scores lowest on its own training players (macro-F1 0.705, against 0.824--0.856 for the others), which suggests that it trained differently. Its training stopped after 7 epochs, so with patience 5 the restored weights come from epoch 2.
Because a single fold can dominate a five-fold mean, Figure~\ref{fig:perfold} shows every fold's result and not just the average.

\textbf{Simulation may be optimistic.}
Our simulator picks activation scales separately for every window, whereas real hardware often fixes them in advance from calibration data.
At 8 bits, simulated W8A8 scored higher than the real full-integer TFLite model in all five folds, by 0.0020 on average (from TFLite fold scores reported to three decimals).
The 4-bit results may be optimistic in the same way, by an amount we did not measure.
Section~\ref{sec:sens} shows how much one such detail can matter: first-layer blocking moves NVFP4's W4A4 change from $-$0.0075 to $-$0.0095.

\textbf{Scope.}
We study one small CNN, one dataset with accelerometer data only, and basketball as a stand-in for hockey.
Test players come from the same games and sessions as the training players, so our results do not cover new teams, games or seasons.
We did no quantization-aware training, which typically recovers part of the low-bit loss.
INT4 and INT8 use coarser activation scales than the FP4 formats, so the INT4 W4A4 result is not a fair format comparison; an INT4 baseline with 16- or 32-value blocks is future work.
Latency was measured on a shared cloud VM (Intel Xeon @ 2.20\,GHz, 2 virtual CPUs; Table~\ref{tab:tflite}), and MXFP4 and NVFP4 have no speed or energy numbers because that hardware has no 4-bit arithmetic.
We plan measurements on a Raspberry Pi and on FP4-capable hardware.

\textbf{Statistics.}
Five folds give limited statistical evidence. They can show a consistent direction, but they cannot establish statistical significance.
A difference that holds in all five folds would occur by chance with probability 0.0625 if a format had no effect, and fold scores are not independent~\cite{bengio2004cv}.
We therefore treat consistent 5-of-5 results as moderate evidence and inconsistent results as no evidence of a difference.

\section{Conclusion}
For a small wrist-worn activity recognizer, FP16 and INT8 cost practically nothing in accuracy.
With 4-bit weights and activations, NVFP4 stays close to FP32 ($-$0.0075 to $-$0.0095 across our simulator variants), while MXFP4 loses 0.0444 to 0.0514 in every fold.
Most of that gap comes from how the round-down scale rule of our MXFP4 baseline handles blocks whose maxima it would clip, not from block size.
Rounding the power-of-two scale up, a conversion recipe the MX specification permits~\cite{rouhani2023mx}, recovers most of the gap at MXFP4's 4.25 bits per weight.
Next, we will apply this pipeline to ice-hockey stick-sensor data, add a block-wise INT4 baseline, and measure the deployable formats on a Raspberry Pi.

\section*{Acknowledgments}
We used Claude (Anthropic), a generative AI assistant, during this work. It assisted with editing the text for grammar and flow, as well as drafting portions of the analysis code (Section~\ref{sec:method}, ``Use of generative AI''). The authors designed the study, reviewed and revised all text, and assume full responsibility for the manuscript's contents. All numerical results reported in the paper originate from the accompanying notebook, executed on Google Colab and hosted on GitHub; an automated validation cell in the notebook verifies each reported value against the program's output.

\bibliographystyle{plainnat}
\bibliography{refs}

@article{hoelzemann2023hangtime,
  author  = {Hoelzemann, Alexander and Romero, Julia Lee and Bock, Marius and Van Laerhoven, Kristof and Lv, Qin},
  title   = {Hang-Time {HAR}: A Benchmark Dataset for Basketball Activity Recognition Using Wrist-Worn Inertial Sensors},
  journal = {Sensors},
  volume  = {23},
  number  = {13},
  pages   = {5879},
  year    = {2023},
  doi     = {10.3390/s23135879}
}

@misc{ocp2023mx,
  author       = {{Open Compute Project}},
  title        = {{OCP} Microscaling Formats ({MX}) Specification, Version 1.0},
  year         = {2023},
  howpublished = {\url{https://www.opencompute.org/documents/ocp-microscaling-formats-mx-v1-0-spec-final-pdf}}
}

@misc{rouhani2023mx,
  author       = {Rouhani, Bita Darvish and Zhao, Ritchie and More, Ankit and Hall, Mathew and Khodamoradi, Alireza and others},
  title        = {Microscaling Data Formats for Deep Learning},
  year         = {2023},
  note         = {arXiv:2310.10537},
}

@misc{nvidia2025nvfp4,
  author       = {Alvarez, Eduardo and Almog, Omri and Chung, Eric and Layton, Simon and Stosic, Dusan and Krashinsky, Ronny and Aubrey, Kyle},
  title        = {Introducing {NVFP4} for Efficient and Accurate Low-Precision Inference},
  year         = {2025},
  month        = jun,
  howpublished = {NVIDIA Technical Blog, \url{https://developer.nvidia.com/blog/introducing-nvfp4-for-efficient-and-accurate-low-precision-inference/}}
}

@misc{micikevicius2022fp8,
  author       = {Micikevicius, Paulius and Stosic, Dusan and Burgess, Neil and Cornea, Marius and Dubey, Pradeep and others},
  title        = {{FP8} Formats for Deep Learning},
  year         = {2022},
  note         = {arXiv:2209.05433},
}

@inproceedings{jacob2018quant,
  author    = {Jacob, Benoit and Kligys, Skirmantas and Chen, Bo and Zhu, Menglong and Tang, Matthew and Howard, Andrew and Adam, Hartwig and Kalenichenko, Dmitry},
  title     = {Quantization and Training of Neural Networks for Efficient Integer-Arithmetic-Only Inference},
  booktitle = {Proceedings of the IEEE/CVF Conference on Computer Vision and Pattern Recognition (CVPR)},
  pages     = {2704--2713},
  year      = {2018},
  doi       = {10.1109/CVPR.2018.00286}
}

@misc{tfliteptq,
  author       = {{TensorFlow}},
  title        = {Post-training quantization},
  howpublished = {TensorFlow Model Optimization documentation, \url{https://www.tensorflow.org/model_optimization/guide/quantization/post_training}},
  year         = {n.d.},
  note         = {Accessed 24 October 2026}
}

@inproceedings{yang2025mxstudy,
  author    = {Yang, Hanmei and Deng, Summer and Nagpal, Amit and Naumov, Maxim and Janani, Mohammad and Liu, Tongping and Guan, Hui},
  title     = {An Empirical Study of Microscaling Formats for Low-Precision {LLM} Training},
  booktitle = {2025 IEEE 32nd Symposium on Computer Arithmetic (ARITH)},
  pages     = {1--8},
  year      = {2025},
  doi       = {10.1109/ARITH64983.2025.00011}
}

@article{daghero2022har,
  author  = {Daghero, Francesco and Burrello, Alessio and Xie, Chen and Castellano, Marco and Gandolfi, Luca and Calimera, Andrea and Macii, Enrico and Poncino, Massimo and Jahier Pagliari, Daniele},
  title   = {Human Activity Recognition on Microcontrollers with Quantized and Adaptive Deep Neural Networks},
  journal = {ACM Transactions on Embedded Computing Systems},
  volume  = {21},
  number  = {4},
  pages   = {1--28},
  year    = {2022},
  doi     = {10.1145/3542819}
}

@article{dehghani2019windows,
  author  = {Dehghani, Akbar and Sarbishei, Omid and Glatard, Tristan and Shihab, Emad},
  title   = {A Quantitative Comparison of Overlapping and Non-Overlapping Sliding Windows for Human Activity Recognition Using Inertial Sensors},
  journal = {Sensors},
  volume  = {19},
  number  = {22},
  pages   = {5026},
  year    = {2019},
  doi     = {10.3390/s19225026}
}

@article{gholamian2020loso,
  author  = {Gholamiangonabadi, Davoud and Kiselov, Nikita and Grolinger, Katarina},
  title   = {Deep Neural Networks for Human Activity Recognition With Wearable Sensors: Leave-One-Subject-Out Cross-Validation for Model Selection},
  journal = {IEEE Access},
  volume  = {8},
  pages   = {133982--133994},
  year    = {2020},
  doi     = {10.1109/ACCESS.2020.3010715}
}

@article{bengio2004cv,
  author  = {Bengio, Yoshua and Grandvalet, Yves},
  title   = {No Unbiased Estimator of the Variance of {K}-Fold Cross-Validation},
  journal = {Journal of Machine Learning Research},
  volume  = {5},
  pages   = {1089--1105},
  year    = {2004}
}

@article{vleugels2021hockey,
  author  = {Vleugels, Robbe and Van Herbruggen, Ben and Fontaine, Jaron and De Poorter, Eli},
  title   = {Ultra-Wideband Indoor Positioning and {IMU}-Based Activity Recognition for Ice Hockey Analytics},
  journal = {Sensors},
  volume  = {21},
  number  = {14},
  pages   = {4650},
  year    = {2021},
  doi     = {10.3390/s21144650}
}

\end{document}